\documentclass{article}
\usepackage[utf8]{inputenc}
\usepackage{cogsys}
\usepackage{natbib}
\usepackage{algorithm}
\usepackage{algpseudocode}
\usepackage{graphicx}
\graphicspath{ {./images/} }
\usepackage[T1]{fontenc}
\usepackage{wrapfig}

\begin{document}

\title{Toward Forgetting-Sensitive Referring Expression Generation for Integrated Robot Architectures}
\author{Tom Williams}{twilliams@mines.edu}
\author{Torin Johnson}{tajohnson@mines.edu}
\author{Will Culpepper}{wculpepper@mines.edu}
\author{Kellyn Larson}{kellynrlarson@gmail.com}
\address{MIRRORLab, Colorado School of Mines, Golden CO, USA}

\date{April 2020}

\vskip 0.2in
\begin{abstract}
 To engage in human-like dialogue, robots require the ability to describe the objects, locations, and people in their environment, a capability known as “Referring Expression Generation.” As speakers repeatedly refer to similar objects, they tend to re-use properties from previous descriptions, in part to help the listener, and in part due to cognitive availability of those properties in working memory (WM). Because different theories of working memory   ``forgetting'' necessarily lead to differences in cognitive availability, we hypothesize that they will similarly result in generation of different referring expressions. To design effective intelligent agents, it is thus necessary to determine how different models of forgetting may be differentially effective at producing natural human-like referring expressions. In this work, we computationalize two candidate models of working memory forgetting within a robot cognitive architecture, and demonstrate how they lead to cognitive availability-based differences in generated referring expressions. 
\end{abstract}

\section{Introduction}






Effective human-robot interaction requires human-like natural language and dialogue capabilities that are sensitive to robots' embodied nature and inherently situated context~\citep{mavridis2015review,tellex2020robots}. In this paper we explore the role that models of Working Memory can play in enabling such capabilities in integrated robot architectures.
While Working Memory has long been understood to be a core feature of human cognition, and thus a central component of cognitive architectures, recent evidence from psychology suggests a conception of working memory that is subtly different from what is implemented in most cognitive architectures. 
Specifically, while most models of working memory in computational cognitive architectures maintain a single central working memory store, Converging evidence from different communities suggests that humans have different resource limitations for different types of information. Moreover, recent psychological evidence suggests that Working Memory may be a limited resource pool, with resources consumed on the basis of the number and type of features retained. This suggests that \textit{forgetting} in Working should be modeled in cognitive architectures as a matter of systematic removal (on the basis of decay or interference) of entity \textit{features} with sensitivity to the resource limitations imposed for the specific \textit{type} of information represented by that feature.
Of course robot cognition need not directly mirror human cognition, and indeed robots have both unique knowledge representational needs and increased flexibility in how resource limitations are implemented. In this work, we present a robot architecture in which (1) independent resource pools are maintained for different robot-oriented types of information; (2)  WM resources are maintained at the feature level rather than entity level; and (3) both interference- and decay-based forgetting procedures may be used. 
This architecture is flexibly configurable both in terms of what type of forgetting procedure is used, and how that model is parameterized. For robot designers, this choice of parameterization may be made in part on the basis of facilitation of interaction. In this paper we specifically consider how the use of different models of forgetting within this architecture lead to different information being retained in working memory, which in turn leads to different referring expressions being generated by the robot, which in turn can produce \textit{interactive alignment} effects purely through Working Memory dynamics.
While in future work it will be important to identify exactly which parameterizations lead to selection of referring expressions that are optimal for effective human-robot interaction and teaming, in this work we take the critical first step of demonstrating, as a proof-of-concept, that decay- and interference-based forgetting mechanisms can be flexibly used within this architecture, and that those policies do indeed produce different natural language generation behavior.

\section{Referring}
\subsection{Models of Referring Expression Generation}
``Referring'' has been referred to as the ``fruit fly''
of language due to the amount of research it has attracted~\citep{van2016computational,gundel2019oxford}. 
In this work, we focus specifically on Referring Expression Generation (REG) \citep{reiter1997building} in which a speaker must choose words or phrases that will allow the listener to uniquely identify the speaker's intended referent. 
REG includes two constituent sub-problems~\citep{gatt2018survey}: referring form selection and referential content determination. While referring form selection (in which the speaker decides whether to use a definite, indefinite, or pronominal form~\citep{poesio2004centering,mccoy1999generating} \citep[see also][]{pal2020cogsci}
) has attracted relatively little attention, referential content determination is one of the most well-explored sub-problems within Natural Language Generation, in part due to the logical nature of the problem that enables it to be studied in isolation, to the point where ``REG'' is typically used to refer to the referential content determination phase alone. In this section we will briefly define and describe the general strategies that have been taken in the computational modelling of referential content determination; for a more complete account we recommend the recent book by \cite{van2016computational}, which provides a comprehensive account of work on this problem.

Referential content determination, typically employed when generating definite descriptions, is the process by which a speaker seeks to determine a set of constraints on known objects that if communicated will distinguish the target referent from other candidate referents in the speaker and listener's shared environment. 
These constraints most often include attributes of the target referent, but can also include relationships that hold between the target and other salient entities that can serve as ``anchors'', as well as attributes of those anchors themselves~\citep{dale1991generating}. 

Three referential content determination models have been particularly influential: the \textit{Full Brevity Algorithm}~\citep{dale1989cooking}, in which the speaker selects the description of minimum length, in order to straightforwardly satisfy Grice's Maxim of Quantity~\citep{grice1975logic}; the \textit{Greedy Algorithm}, in which the speaker incrementally adds to their description whatever property rules out the largest number of distractors\citep{dale1992generating}; and the \textit{Incremental Algorithm (IA)}, in which the speaker incrementally adds properties to their description in order of \textit{preference} so long as they help to rule out distractors~\citep{dale1995computational}. A key aspect of the IA is its ability to \textit{overspecify} through its inclusion of properties that are not strictly needed from a logical perspective to single out the target referent, but are nevertheless included due to being highly preferred; a behavior also observed in human speakers~\citep{engelhardt2006speakers}. 

Because the IA's behavior is highly sensitive to preference ordering~\citep{gatt2007evaluating}, there has been substantial research seeking to determine what properties are in general psycholinguistically preferred over others~\citep{belke2002tracking}, or to automatically learn optimal preference orderings~\citep{koolen2012learning}. As highlighted by \cite{goudbeek2012alignment}, however, this focus on a uniform concept of ``preference'' obscures a much more complex story that relates to fundamental debates over the extent to which speakers leverage listener knowledge during sentence production. A notion of ``preference'' as encoded in the IA could be egocentrically grounded, with speakers ``prefer'' concepts that are easy for themselves to assess or cognitively available to themselves \citep{keysar1998egocentric}; it could be allocentrically grounded, with speakers intentionally seeking to facilitate the listener's ease of processing~\citep{janarthanam2009learning}; or a hybrid model could be used, in which egocentric and allocentric processes compete~\citep{bard2004referential}, with egocentrism vs. allocentrism ``winning out'' on the basis of factors such as cognitive load~\citep{fukumura2012producing}.
These approaches, which require accounting for listener knowledge to be slow and intentional, stand in contrast to memory-oriented account of referential content determination in which such accounting can naturally occur as a result of priming.

\subsection{Memory-Oriented Models of Referring Expression Generation}

\citet{pickering2004toward}'s \textit{Interactive Alignment} model of dialogue suggests that dialogue is a highly negotiated process (see also \cite{clark1986referring}) in which priming mechanisms lead interlocutors to influence each others' linguistic choices at the phonetic, lexical, syntactic and semantic levels, through mutual activation of phonetic, lexical, syntactic, and semantic structures and mental representations, as in the case of lexical entrainment~\citep{brennan1996conceptual}.

While there has been extensive evidence for lexical and syntactic priming, research on semantic or conceptual priming in dialogue has only relatively recently become a target of substantial investigation~\citep{gatt2011attribute}. A theory of dialogue including semantic or conceptual priming would suggest that the properties or attributes that speakers choose to highlight in their referring expressions (e.g., when a speaker chooses to refer to an object as   ``the large red ball'' rather than   ``the sphere'') should be due in part to these priming effects. And in fact, as demonstrated by \citet{goudbeek2010preferences}, speakers can in fact be influenced through priming to use attributes in their referring expressions that would otherwise have been dispreferred. 

These findings have motivated dual-route computational models of dialogue~\citep{gatt2011attribute,goudbeek2011referring} in which the properties used for referring expression selection are made on the basis of interaction between two parallel processes, each of which is periodically called upon to provide attributes of the target referent to be placed into a WM buffer that is consulted when properties are needed for RE generation (at which point selected properties are removed from that buffer). 
The first of these processes is a priming-based procedure in which incoming descriptions trigger changes in activation within a spreading activation model, and properties are selected if they are the highest-activation properties (above a certain threshold) for the target referent.
The second of these processes is a preference-based procedure in which a set of properties is generated by a classic REG algorithm \citep[cp.][]{gatt2018survey} such as the Incremental Algorithm, in which properties are incrementally selected according to a pre-established preference ordering designed or learned to reflect frequency of use, ease of cognitive or perceptual assessability, or some other informative metric~\citep{dale1995computational}.

One advantage of this type of dual process model is that it accounts for audience design effects (in which speaker utterances at least appear to be intentionally crafted for ease-of-comprehension) within an egocentric framework, 
by demonstrating how priming influences on WM can themselves account for listener-friendly referring expressions. That is, if a concept is primed by an interlocutor's utterance, a speaker will be more likely to use that concept in their own references simply because it is in WM, with the side effect that that property will then be easy to process by the interlocutor responsible for its' inclusion in WM in the first place~\citep{vogels2015cognitive}. Moreover, this phenomenon aligns well with evidence suggesting that despite the prevalence of lexical entrainment and alignment effects, people are actually slow to explicitly take others' perspectives into account~\citep{bard2000controlling,fukumura2012producing,gann2014speaking}.

Another advantage of this type of dual process model is its alignment with overarching dual-process theories of cognition \citep[e.g.,][]{kahneman2011thinking,evans2013dual,sloman1996empirical,oberauer2009design}: the first priming-driven process for populating WM, grounded in semantic activation, can be viewed as a reflexive System One process, whereas the second preference-driven process leveraging the Incremental Algorithm can be viewed as a deliberative System Two process. Of course in the model under discussion the two routes do not truly compete with each other or operate on different time courses, but are instead essentially sampled between; however, it is straightforward to imagine how the two processes used in this type of model could be instead deployed in parallel.

One major disadvantage of this type of model, however, is that its focus with respect to WM is entirely on retrieval (i.e., how priming and preference-based considerations impact what information is retrieved from long-term memory into WM), and fails to satisfactorily account for maintenance within WM. Within \citet{gatt2011attribute}'s model, as soon as a property stored in WM is used in a description, it is removed from WM so that that space is available for another property to be considered. This behavior seems counter-intuitive, as it ensures that representations are removed from WM at just the time when it is established to be important and useful, which should be a cue to retain said representations rather than dispose of them. 

Moreover, this model is surprisingly organized from the perspective of models of WM such as \citet{cowan2001magical}'s, in which WM is comprised of the set of all activated representations, of which a small subset (e.g., three or four) are maintained in the focus of attention. In \citet{gatt2011attribute}'s model, in contrast, activated representations are used as just one source populating WM, and decaying activation within the spreading activation network results in representations losing activated status without also being removed from WM. This suggests that the WM buffer within \citet{gatt2011attribute}'s model may in fact be better conceptualized as a model of the focus of attention (an interpretation also justified by the two-item size limitation of their WM buffer) than as a model of WM.

A final complication for this model is its speaker-blind handling of priming. Specifically, within \citet{gatt2011attribute}'s model a speaker's utterances are only primed by their interlocutor's utterances, when in fact the choices a speaker makes should also impact the choices they themselves make in the future~\citep{shintel2007you}, either due to Gricean notions of cooperativity~\citep{grice1975logic} or, as we argue, because a speaker's decision to refer to a referent using a particular attribute should make that attribute more cognitively available to themselves in the immediate future.

These concerns are addressed by our previously proposed model of robotic short-term memory \citep{williams2018icsr}, in which speakers rely on the contents of WM for initial attribute selection and, if their selected referring expression is not fully discriminating, select additional properties using a variant of the IA. While this does not align with dual-process models of cognition, it does account for both encoding and maintenance of WM, and provides a potentially more cognitively plausible account of REG with respect to WM dynamics. One shortcoming shared by both models, however, is neither the dual-process model of \citet{gatt2011attribute} nor our WM-focused model appropriately account for 
when and how information is removed from WM over time, or how this impacts REG.

Different theories of WM   ``forgetting'' necessarily lead to predicted differences in cognitive availability. Accordingly, these different models of forgetting should similarly predict cognitive availability-based differences in the properties selected during REG. 
To design effective intelligent agents, it is thus necessary to determine how different models of forgetting may be differentially effective at producing natural human-like referring expressions. 

In this work, we first computationalize two candidate models of WM forgetting within a robot cognitive architecture. Next, we propose a model of REG that is sensitive to the WM dynamics of encoding, retrieval, maintenance, \textit{and forgetting}, discuss the particulars of deploying this type of model within an integrated robot architecture, where WM resources are divided by domain (i.e., people, locations, and objects) rather than by modality (i.e., visual vs. verbal). Finally, we provide a proof-of-concept demonstration of two parametrizations of our model into an integrated robot cognitive architecture, and demonstrate how these different parametrizations lead to cognitive availability-based differences in generated referring expressions.


\section{Models of Forgetting in Working Memory}

Models of forgetting in Working Memory are typically divided into two broad categories~\citep{reitman1971mechanisms,jonides2008mind,ricker2016decay}: decay-based models, and interference-based models. 

\subsection{Decay-Based Models}

Decay-based 
models of WM~ \citep{brown1958some,atkinson1968human}, that 
time plays a causal role in WM forgetting, with 
a representation's inclusion in WM made on the basis of a level of activation that gradually decays over time if the represented information is not used or rehearsed. Accordingly,  in such models, a piece of information is   ``forgotten'' with respect to WM if it falls below some threshold of activation due to disuse. This model of forgetting is intuitively appealing due to the clear evidence that memory performance decreases over time~\citep{brown1958some,kep,ricker2016decay}.\\

\noindent\textbf{Computational Advantages and Limitations:}
As with Gatt et al., spreading activation networks can be used to elegantly model how activation of representations impacts the rise and fall of activation of semantically related pieces of information. 
One disadvantage of this approach, however, is that activation levels need to be continuously re-computed for each knowledge representation in memory. While this may be an accurate representation of actual cognitive processing, artificial cognitive systems do not enjoy the massively parallelized architectures enjoyed by biological cognitive systems, meaning that this approach may face severe scaling limitations in practice.\\ 

\noindent\textbf{Computational Model:}
To allow for straightforward comparison with other models of forgetting, we define a simple model of decay that operates on individual representations outside the context of a semantic activation network. We begin by representing WM as a set $WM = {Q_0,\dots,Q_n}$, where $Q_i$ is a mental representation of a single entity, represented as a queue of properties currently activated for that entity.
Next, we define an encoding procedure that specifies how the representations in WM are created and manipulated on the basis of referring expressions generated either by the agent or its' interlocutors.  As shown in Alg.~\ref{alg:encoding}, this procedure operates by considering each property included in the referring expression, and updating the queue used to represent the entity being described, by placing that property in the back of the queue, or by moving the property to the back of the queue if it's already included in the representation.
Note that this procedure can be used either after each utterance is heard (in which case the representation is updated based on all properties used to describe the entity) or incrementally (in which case the representation is updated after each property is heard). If used incrementally, then forgetting procedures may be interleaved with representation updates.
Finally, we define a model of decay that operates on these representations. 
As shown in Alg.~\ref{alg:decay}, this procedure operates by removing the property at the front of queue $Q$ at fixed intervals defined by decay parameter $\delta$.

\begin{algorithm}
    \caption{Per-entity encoding model}\label{alg:encoding}        
    \begin{algorithmic}[1]
    \Procedure{Encode}{$R,P,Q$}
            \State{$R$: the object being described}
            \State{$P$: the set of properties being used by the speaker or hearer to describe $R$}
            \State{$Q$: set of per-entity property queues.} 
    \ForAll{$p\in P$}
        \State $Q[R] = Q[R] \setminus p$
        \State{$push(Q[R],p)$}
    \EndFor
    \EndProcedure
    \end{algorithmic}
\end{algorithm}

\begin{algorithm}
\begin{algorithmic}
    \caption{Per-entity decay model.}\label{alg:decay}
    \Procedure{PeriodicDecay}{$Q,\delta$}
    \State $Q$: the per-entity property queue
    \State $\delta$: decay parameter
    \Repeat 
    \Comment Every $\delta$ seconds
    \State $pop(Q)$
    \Until $Q=\emptyset$
    \EndProcedure
    \end{algorithmic}
\end{algorithm}

\subsection{Interference-Based Models}

In contrast, interference-based models~\citep{waugh1965primary} argue that WM is a fixed-size buffer 
in which
a piece of information is  ``forgotten'' with respect to WM if it is removed (due to, e.g., being the least recently used representation in WM memory) to make room for some new representation.

Interference-Based Models have been popular for nearly as long as decay-based models~\citep{keppel1962proactive} due to observations that the evidence used as evidence for ``decay'' can just as easily be used as evidence for forgetting due to intra-representational interference, as longer periods of time directly correlate with higher frequencies of interfering events~\citep{lewandowsky2015rehearsal,oberauer2008forgetting}, and because tasks with varying temporal lengths but consistent overall levels of interference have been shown to yield similar rates of forgetting, thus failing to support time-based 
decay~\citep{oberauer2008forgetting}.
Recent work has trended towards interference-based accounts of forgetting, with a number of further debates and competing models opening up within this broad theoretical ground.

First, there is debate as to whether interference \textit{alone} is sufficient to explain forgetting, or whether time-based decay still plays some role in conjunction with interference. Recent work suggests that in fact these two models may be differentially employed for different types of representations, with phonological representations forgotten due to interference and non-phonological representations forgotten due to a combination of interference and time-based decay~\citep{ricker2016decay}.

Second, within interference-based models, there exist competing models based on reasons for displacement. In particular, while theories of pure displacement~\citep{waugh1965primary} posit that incoming representations replace maintained representations on the basis of frequency or recency of use, or on the basis of random chance (similar to caching strategies from computing systems research~\citep{press2014caching}), theories of retroactive interference instead posit that replacements are made on the basis of semantic similarity, with representations "forgotten" if they are too similar to incoming representations~\citep{wickelgren1965acoustic,lewis1996interference}.

Third, within both varieties of interference-based models, there has been recent debate on the structure and organization of the capacity-limited medium of WM. \cite{ma2014changing} contrasts four such models: (1) slot models, in which a fixed number of slots are available for storing coherent representations~\citep{miller1956magical,cowan2001magical}; (2) resource models, in which a fixed amount of representational medium can be shared between an unbounded number of representations (with storage of additional features in one representation resulting in fewer feature-representing medium being available for other representations)~\citep{wilken2004detection}; (3) discrete-representation models, in which a fixed number of feature ``quanta'' are available to distribute across representations~\citep{zhang2008discrete}; and (4) variable-precision models, in which working memory precision is statistically distributed~\citep{fougnie2012variability}.\\

\noindent\textbf{Computational Advantages and Limitations:}
One advantage of interference-based models for artificial cognitive agents is decreased computational expense, as only a fixed number of entities or features must be maintained in WM, and WM need not be updated at continuous intervals if no new stimuli are processed. Rather, WM only needs to be updated when (1) new representations are encoded into WM, or (2) existing representations are manipulated. Another advantage of this approach is its conceptual alignment with the process of \textit{caching} from computer science, which means that caching mechanisms from computing systems research, such as \textit{least-recently-used} and \textit{least-frequently-used} caching policies, can be straightforwardly leveraged, with prior work providing substantial information about their theoretical properties and guarantees. In fact, recent work has explored precisely how caching strategies from computer science can be used for this purpose~\citep{press2014caching}). 
Within the interference-based family of models, slot-based and discrete-representation models are likely the most easy to computationalize due to the ephemeral and undiscretized nature of ``representational medium.''\\
    
\noindent\textbf{Computational Model:}
%
%
To model interference-based forgetting, we use the same WM representation and encoding procedure as used to model decay-based forgetting, and propose a new model designed specifically for robotic knowledge representation. This model can be characterized as a per-entity discrete-representation displacement model.
As shown in Alg.~\ref{alg:displacement}, this procedure operates by removing properties at the front of queue \textit{Q} whenever the size of $Q$ is greater than some size limitation imposed by parameter $\alpha$. 
This model is characterized as discrete-representation because resource limitations are imposed at the level of discrete features rather than holistic representations. It is characterized as a displacement model because features are replaced on the basis of a Least-Recently Used (LRU) caching strategy~\citep{knuth1997art} rather than on the basis of semantic similarity, due to the pragmatic difficulty of assessing the similarity of different categories of properties without mandating the use of architectural features such as well-specified conceptual ontologies~\citep[cp.][]{tenorth2009knowrob,lemaignan2010oro}, which may not be available in all robot architectures.
This model is characterized as per-entity because resource limitations are imposed locally (i.e., for each entity) rather than globally (i.e., shared across all entities). While this is obviously not a cognitively plausible characteristic, it was selected, as a starting point, to reduce the need for coordination across architectural components and to facilitate improved performance (as entity representations need not compete with each other). However, if desired, it would be straightforward to extend this approach to allow for imposition of global resource constraints.



\begin{algorithm}
\begin{algorithmic}
    \caption{Per-entity discrete representation displacement model.}\label{alg:displacement}
    \Procedure{Displacement}{$Q,\alpha$}
    \State $Q$: the per-entity property queue
    \State $\alpha$: maximum buffer size
    \Loop 
    \If{$|Q| > \alpha$}
    \State $pop(Q)$
    \EndIf
    \EndLoop
    \EndProcedure
    \end{algorithmic}
\end{algorithm}








\section{Models of Working Memory in Integrated Robot Architectures}

Memory modeling has long been a core component of cognitive architectures, due to its central role in cognition~\citep{baxter2011memory}. 
The relative attention paid to WM has, however, varied widely between cognitive architectures. ARCADIA, for example, has placed far more emphasis on attention than on memory~\citep{bridewell2016theory}. While ARCADIA does include a visual short term memory component, it is treated as a ``potentially infinite storehouse,'' with consideration of resource constraints left to future work \citep{bridewell2016theory}. 

ACT-R and Soar, in contrast, do place larger emphases on WM. 
ACT-R did not originally have explicit WM buffers, instead  implicitly representing WM as the subset of LTM with activation above some particular level~\citep{anderson1996working}, with ``forgetting'' thus modeled through activation decay~\citep[cp.][]{cowan2001magical}. In more recent incarnations of ACT-R, a very small short-term buffer is maintained, with contents retrieved from LTM on the basis of both base-level activation (on the basis of recency and frequency) and informative cues.
Similarly, Soar~\cite{laird2012soar} has long emphasized the role of WM, due to its central focus on problem solving through continuous manipulation of WM~\citep{rosenbloom1991preliminary}. And while Soar did not initially represent WM resource limitations~\citep{young1999soar}, it has also by now long included decay-based mechanisms on at least a subset of WM contents~\citep{chong2003addition,nuxoll2004comprehensive}, as well as base-activation and cue-based retrieval methods such as those mentioned above~\cite{jones2016efficient}.
While the models may be intended \textit{primarily} as models of human cognition, flaws and all, rather than as systems for enabling effective task performance through whatever means necessary regardless of cognitive plausibility, a significant body of work has well demonstrated the utility of these architectures for cognitive robotics~\citep{laird2012cognitive,laird2017standard,kurup2012can} and human-robot interaction~\cite{trafton2013act}.

There has also been significant work in robotics seeking to use insights from psychological research on WM to better inform the design of select components of larger robot cognitive architectures that do not necessarily aspire towards cognitive plausibility. For example, a diverse body of researchers has collaborated on the development and use of the WM Toolkit~\citep{phillips2005biologically,gordon2006system,kawamura2008implementation,persiani2018working}; a software toolkit that maintains pointers to a fixed number of chunks containing arbitrary information. At each timestep, this toolkit proposes a new set of chunks, and then uses neural networks to select a subset of these chunks to retain. This model has been primarily used for enabling \textit{cognitive control}, in which the link between robot perception and action is modulated by a learned memory management policy. 
 
Models of WM have also been leveraged within the field of Human-Robot Interaction. \cite{broz2012interaction}, for example, specifically model the episodic buffer sub-component of WM~\citep{baddeley2000episodic}. 
\cite{baxter2013cognitive} 
leverages models of WM to better enable non-linguistic social interaction through alignment, similar to our approach in this work. Researchers have also leveraged models of WM to facilitate communication. For example, \cite{hawes2007towards} leverage a model of WM within the CoSy architecture, with concessions made to accommodate the realistic needs of integrated robot architectures, in which specialized representations are stored and manipulated within distributed, parallelized, domain-specific components~\cite[see also][]{williams2016aaai}. Similarly, in our own work within the DIARC architecture~\citep{scheutz2019overview}, we have demonstrated (as we will further discuss in this paper) the use of distributed WM buffers associated with such architectural components~\citep{williams2018icsr}, as well as hierarchical models of common ground \citep{williams2019oxford} jointly inspired by models of WM~\citep[e.g.][]{cowan2001magical} and models of \textit{givenness} from natural language pragmatics~\citep[e.g.][]{gundel1993cognitive}. In the next section we propose a new architecture that builds on this previous work of ours to allow for flexible selection (and comparison between) between different models of forgetting.

\section{Proposed Architecture}
Our forgetting models were 
integrated into 
the Distributed, Integrated, Affect, Reflection, Cognition (DIARC) Robot Cognitive architecture: a component-based architecture designed with a focus on robust spoken language understanding and generation~\citep{scheutz2019overview}. DIARC's Mnemonic and Linguistic components integrate via a \textit{consultant framework} in which different architectural components (e.g., vision, mapping, social cognition) serve as heterogeneous knowledge sources that comprise a distributed model of Long Term Memory~\citep{williams2017iib,williams2016aaai}. 

These consultants are used by DIARC's natural language pipeline for the purposes of reference resolution and REG. 
In recent work we have extended this framework to produce a new \textit{Short-Term Memory Augmented} consultant framework in which consultants additionally maintain, for some subset of the entities for which they are responsible, a short term memory buffer of properties that have recently been used by interlocutors to refer to those entities. In this work, we build upon that STM (Short Term Memory)-Augmented Consultant Framework through the introduction of a new architectural component, the \textsc{WM Manager}, which is responsible for implementing the two forgetting strategies introduced in the previous section. 

Our model aligns with two key psychological insights. First, converging evidence from different communities suggests that humans have different resource limitations for different types of information~\citep{wickens2008multiple}, due to either
decreased interference between disparate representations~\citep{oberauer2012modeling} or 
the use of independent domain-specific resource pools~\citep{baddeley1992working,logie1995visuo}. Our approach takes a robot-oriented  perspective on this second hypothesis, with our use of the WM-Augmented consultant framework resulting in independent resource pools maintained for different types of entities (e.g., objects vs. locations vs. people) rather than for different modalities (e.g., visual vs. auditory) or different codes of processing (e.g., spatial vs. verbal).

Second, while early models of WM suggested that WM resource limitations are bounded to a limited number of chunks~\citep{miller1956magical}, more recent models instead suggest that the size of WM is affected by the complexity of those chunks~\citep{mathy2012s}, and that maintaining multiple features of a single entity may detract from the total number of maintainable entities, and accordingly, the number of features maintainable for other entities~\citep{alvarez2004capacity,oberauer2016limits,taylor2017does}. Our approach, again, takes a robot-oriented perspective on these models, maintaining WM resources at the feature-level rather than entity-level, while enabling additional flexibility that may not be reflected in human cognition. Specifically, instead of enforcing global resource limits, we allow for flexible selection between decay-based and interference-based (i.e., resource-limited) memory management models, as well as for simultaneous employment of both models, in order to model joint impacts of interference and decay as discussed by~\cite{ricker2016decay}. Moreover, while we currently focus on local (per-entity) feature-based resource limitations, our system is designed to allow for global resource limitations~\citep[cp.][]{just1992capacity,ma2014changing} in future work, due to our use of a global \textsc{WM Manager} Component.

Using this architecture, our different forgetting models can differentially affect REG without any direct interaction with the REG process. Rather, the \textsc{WM Manager} simply interfaces with the WM buffers maintained by each distributed consultant, which then implicitly impacts the properties available to the SD-PIA algorithm that we use for REG. This algorithm takes a lazy approach to REG in which the speaker initially attempts to rely only on the properties that are currently cached  in WM, and only if this is insufficient to distinguish the target from distractors does the speaker retrieve additional properties from long-term memory.


\begin{figure}
    \centering
    \includegraphics[ 
    width=\linewidth]{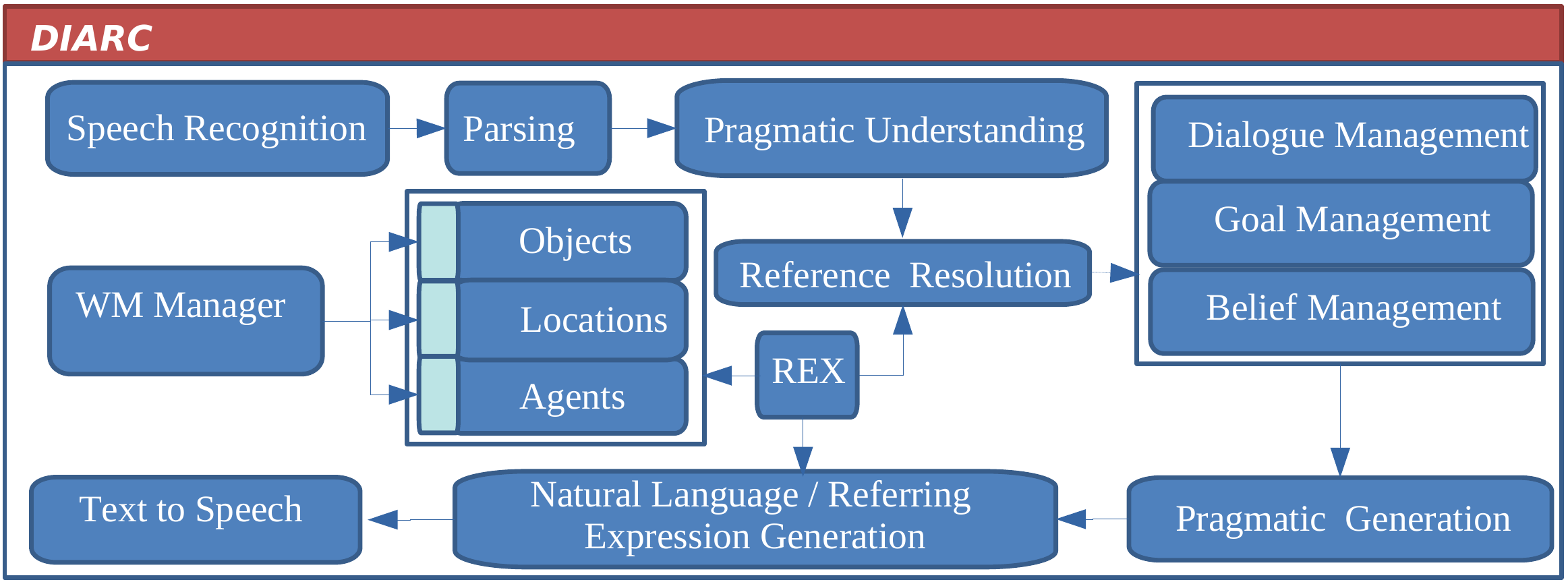}
    \caption{Architecture Diagram}
    \label{fig:architecture}
\end{figure}

Incorporating the \textsc{WM Manager} into DIARC yields the configuration shown in Fig.~\ref{fig:architecture}. When a teammate speaks to a robot running this configuration, text is recognized and then 
semantically parsed 
by the TLDL Parser~\citep{scheutz2017spoken}, after which intention inference is performed
by the Pragmatic Understanding Component~\citep{williams2015aaai}, whose results are provided to the Reference Resolution module of the Referential Executive (REX), which leverages the
%
GROWLER 
algorithm~\citep{williams2018mrhrc:growler} (see also~\citep{williams2016hri,williams2019oxford}), which 
searches through Givenness Hierarchy (GH)~\citep{gundel1993cognitive} informed data structures (Focus, Activated, Familiar) representing a hierarchically organized cache of relevant entity representations. This is important to highlight due to its relation to the WM buffers described in this work. While the WM buffers described in this work serve as a model of the robot's own WM, the Referential Executive's data structures can instead be viewed as either second-order theory-of-mind data structures
or as a form of hierarchical common ground.

When particular entities are mentioned by the robot's interlocutor or by the robot itself, pointers to the full representations for these entities (located in the robot's distributed long-term memory) are placed into the Referential Executive's GH-informed data structures, and the properties used to describe them are placed into the robot's distributed WM buffers. In addition, properties are placed into WM whenever the robot affirms that those properties hold through an LTM query. Critically, this can happen when considering other entities. For example, if property $P$ is used during reference resolution, then $P$ will be placed into WM for any distractors for which $P$ is affirmed to hold, before being ruled out for other reasons. Similarly, if $R$ is the target referent during REG, and if property $P$ holds for $R$ and is considered for inclusion in the generated RE, then for any distractors for which $P$ also holds, $P$ will be added to those distractors' WM buffers at the point where it is affirmed that $P$ cannot be used to rule out those distractors because it holds for them as well.
Once Reference Resolution is completed, 
if the robot decides to respond to the human utterance, it does so through a process that is largely the inverse of language understanding, including a dedicated
REG  module, which uses the properties cached in WM, along with properties retrievable from Long-Term Memory, to translate this intention into text~\citep{williams2017inlg,williams2018icsr}.
\section{Experimental Setup and Results}



\begin{table}[h]
    \centering
    \begin{tabular}{c|c|c|c|c|c|c}
        Turn & \multicolumn{2}{c|}{Human's Turn} & \multicolumn{4}{c}{Robot's Turn}\\\hline
          & Face & Description & Face & \multicolumn{3}{c}{Description}\\\hline
          &  &  &  & Decay & Interference & No WM\\\hline\hline
         1 & 1 & $H_L$, $G_M$, $C_L$, $G_Y$ & 1 & $H_L$, $C_L$, $G_Y$ & $H_L$, $C_L$, $G_Y$  & $H_L$, $C_L$, $G_Y$\\\hline
         2 & 2 & $G_F$, $G_Y$ & 5 & $H_D$, $G_M$ & $H_D$, $G_M$  & $H_D$, $G_M$\\\hline
         3 & 3 & $G_F$, $C_H$, $G_N$  & 2 & $G_F$, $G_Y$ & $G_F$, $G_Y$  & $G_F$, $G_Y$\\\hline
         4 & 4 & $H_L$, $H_S$, $C_L$, $G_N$  &1 & $H_S$, $C_L$, $G_Y$ & $H_L$, $C_L$, $G_Y$ & $H_L$, $C_L$, $G_Y$\\\hline
         5 & 5 & $H_D$, $H_S$, $C_T$, $G_M$ & 3 & $G_F$, $C_H$ &  $G_F$, $C_H$ & $G_F$, $C_H$\\\hline
         6 & 6 & $H_S$, $C_H$, $G_Y$ & 1 & $C_L$, $G_Y$ & $H_S$, $C_L$, $G_Y$ & $H_L$, $C_L$, $G_Y$\\\hline
    \end{tabular}

    \caption{6 Face Case Study. Column contents denote properties used to describe target under each forgetting model, listed in the conssitent order below for easy comparison rather than in order selected. Properties: $H_L$ = \textsc{light-hair(X)}, $H_D$ = \textsc{dark-hair(X)}, $H_S$ = \textsc{short-hair(X)}, $G_M$ = \textsc{male(X)}, $G_F$ = \textsc{female(X)}, $C_T$ = \textsc{t-shirt(X)}, $C_L$ = \textsc{lab-coat(X)}, $C_H$ = \textsc{hoodie(X)}, $G_Y$ = \textsc{glasses(X)}, $G_N$ = \textsc{no-glasses(X)}. }
    \label{tab:casestudies}
\end{table}

We analyzed the proposed architecture by assessing two claims: (1) The proposed architecture demonstrates interactive alignment effects purely through WM dynamics; and (2) the proposed architecture demonstrates different referring behaviors when different models of forgetting are selected.
These claims were assessed within the context of a ``Guess Who''-style game in which partners take turns describing candidate referents (assigned from a set of 16 faces). 
On each player's turn, they are assigned a referent, and must describe that referent using a referring expression they believe will allow their interlocutor to successfully identify it (a process of REG). The other player must then process their interlocutor's referring expression and identify which candidate referent they believe to be their interlocutor's target (a process of Reference Resolution).




Ideally, we would have assessed our claims in a setting in which a robot played this reference game with a naive human subject. This proved to be impossible due to the COVID-19 global pandemic. Instead, we present a case study in which a series of three six-round reference games are played between robot agents and a single human agent. All three games followed the same predetermined order of candidate referents and used the same pre-determined human utterances. The robot's responses were generated autonomously, with the robot in each of the three games using a different model of forgetting. In the first game, the robot uses our decay-based model of forgetting with $\delta=10$; in the second game, the robot uses our interference-based model of forgetting with $\alpha=2$; in the third game, the robot did retain any properties in short-term memory at all. 

The referring behavior under each model of forgetting is
shown in Tab.~\ref{tab:casestudies}. As shown in this table, the three examined models perform similarly in initial dialogue turns, but increasingly diverge over time. To help explain the observed differences in robot behavior, we examine specifically turn 6, in which the robot refers to Face 1 for the third time. This face could ostensibly be referred to using four properties: \textsc{light-hair(X)}, \textsc{short hair(X)}, \textsc{lab-coat(X)}, and \textsc{glasses(X)}. 

The architectural configuration that did not maintain representations in WM (No WM) operated according to the DIST-PIA algorithm~\citep{williams2017inlg}, which is a version of the Incremental Algorithm that is sensitive to uncertainty and that allows for distributed sources of knowledge. This algorithm first considers the highly preferred property \textsc{light-hair(X)}, which is selected because it applies to Face 1 while ruling out distractors. Next, it considers \textsc{short-hair(X)}, which it ignores because while it applies to Face 1, the faces with short hair are a subset of those with light hair, and thus \textsc{short-hair(X)} is not additionally discriminative. Next, the algorithm considers \textsc{lab-coat(X)}, which it selects because it applies to Face 1 and rules out further distractors. Finally, to complete disambiguation, the algorithm considers and selects \textsc{glasses(X)}.

In contrast, the configuration that used the decay model had the following properties in WM: \{\textsc{lab-coat(X)}, \textsc{short-hair(X)}, \textsc{glasses(X)}\} (ordered from least-recently used to most-recently used\footnote{Future work should consider other algorithmic configurations, such as having properties within WM considered in the reverse order, or according to the preference ordering specified by the target referent's consultant.}). The algorithm starts by considering the properties stored in WM, beginning with \textsc{lab-coat(X)}, which is selected because it applies to Face 1 while ruling out distractors. Next, it considers \textsc{short-hair(X)}, which is ignored because the set of entities with short hair is a subset of those wearing lab coats, and thus this is not additionally discriminative. Next, it considers \textsc{glasses(X)}, which it selects because it applies to Face 1 and rules out distractors. In fact, \{\textsc{lab-coat(X)}, \textsc{glasses(X)}\} is fully discriminating for Face 1, so no further properties are needed.

Finally, the configuration that used the interference model had the following properties in WM: \{\textsc{short-hair(X)}\, \textsc{glasses(X)}\}. This is easy to see as those properties were recently used in the Human's description of Face 6, and thus would have been considered for Face 1 when ruling it out during reference resolution. The algorithm thus starts by considering both of these properties, which are both selected because they apply to Face 1 and rule out distractors. However, because these are not sufficient for full disambiguation, the algorithm must also retrieve another property from LTM, i.e., \textsc{lab-coat(X)}, which allows for completion of disambiguation. 

The differences in behavior demonstrated in this simple example validate both our claims. First, the proposed architecture's ability to demonstrate interactive alignment effects purely through WM dynamics is demonstrated by the systems' tendency to re-use properties originating from its interlocutor.
Second, this example clearly demonstrates that the proposed architecture demonstrates different referring behaviors when different models of forgetting are selected.

\section{Conclusions and Future Work}

We have presented a flexible set of forgetting mechanisms for integrated cognitive architectures, and conducted a preliminary, proof-of-concept demonstration of these mechanisms, showing that they lead to different referring expressions being generated due to differences in cognitive availability between different properties. The next step of our work will be to fully explore the implications of different parametrizations of each of our presented mechanism, as well as the combined use of these mechanisms, on REG, and whether the referring expressions generated under different parametrizations are comparatively more or less natural, human-like, or effective, which would present obvious benefits 
%
%
for interactive, intelligent robots. 
In addition, 
the perspective taken in this paper may also yield insights and benefits for cognitive science more broadly. Specifically, we argue that our perspective may suggest alternative interpretations of the role of cognitive load on attribute choice. In \citet{goudbeek2011referring}'s work building on \citet{gatt2011attribute}'s model, they suggest that when speakers are under high cognitive load, they rely less on previously primed attributes, and are thus more likely to rely on their stable preference orderings. Their explanation for this finding is that a decrease in available WM capacity leads to an inability to retrieve dialogue context into WM. We suggest an alternative explanation: cognitive load leads to decreased priming not because priming-activated representations cannot be retrieved into WM, but because those representations are less likely to be in WM in the first place. 
An additional promising direction for future work will thus be to compare the ability of the model presented in this paper to those presented by \citet{goudbeek2011referring} with respect to 
modeling of REG under cognitive load.

\setlength{\bibsep}{0pt plus 0.3ex}
\bibliographystyle{cogsysapa}
\bibliography{references}

\end{document}